\documentclass[10pt,twocolumn,letterpaper]{article}

\usepackage{cvpr}
\usepackage{times}
\usepackage{epsfig}
\usepackage{graphicx}
\usepackage{amsmath}
\usepackage{amssymb}

\usepackage{booktabs} 
\usepackage{mathrsfs}
\usepackage{multirow}
\usepackage{graphics}
\usepackage{subfigure}
\usepackage{psfrag}
\usepackage{stfloats}
\usepackage{algorithm}
\usepackage{algorithmic}
\usepackage{stfloats}
\usepackage{amsfonts}
\usepackage{color}
\usepackage[normalem]{ulem}

\usepackage{booktabs}
\usepackage{slashbox}

\def\colspaceS{2.25mm}
\def\colspaceM{4.0mm}

\def\colspaceDD{9.25mm}

\usepackage[breaklinks=true,bookmarks=false]{hyperref}

\cvprfinalcopy 

\def\cvprPaperID{1342} 

\ifcvprfinal\fi
\begin{document}

\title{Memory Matching Networks for One-Shot Image Recognition\thanks{{\small This work was performed at Microsoft Research Asia.}}}

\author{Qi Cai $^{\dag}$, Yingwei Pan $^{\dag}$, Ting Yao $^{\ddag}$, Chenggang Yan $^{\S}$, and Tao Mei $^{\ddag}$ \\
{\centering$^{\dag}$ University of Science and Technology of China, Hefei, China}\\
{\centering$^{\ddag}$ Microsoft Research, Beijing, China}\\
{\centering$^{\S}$ Hangzhou Dianzi University, Hangzhou, China}\\
{\tt\small \{cqcaiqi, panyw.ustc\}@gmail.com, \{tiyao, tmei\}@microsoft.com, cgyan@hdu.edu.cn}
}

%

\maketitle
\thispagestyle{empty}

\begin{abstract}
In this paper, we introduce the new ideas of augmenting Convolutional Neural Networks (CNNs) with Memory and learning to learn the network parameters for the unlabelled images on the fly in one-shot learning. Specifically, we present Memory Matching Networks (MM-Net) --- a novel deep architecture that explores the training procedure, following the philosophy that training and test conditions must match. Technically, MM-Net writes the features of a set of labelled images (support set) into memory and reads from memory when performing inference to holistically leverage the knowledge in the set. Meanwhile, a Contextual Learner employs the memory slots in a sequential manner to predict the parameters of CNNs for unlabelled images. The whole architecture is trained by once showing only a few examples per class and switching the learning from minibatch to minibatch, which is tailored for one-shot learning when presented with a few examples of new categories at test time. Unlike the conventional one-shot learning approaches, our MM-Net could output one unified model irrespective of the number of shots and categories. Extensive experiments are conducted on two public datasets, i.e., Omniglot and \emph{mini}ImageNet, and superior results are reported when compared to state-of-the-art approaches. More remarkably, our MM-Net improves one-shot accuracy on Omniglot from 98.95\% to 99.28\% and from 49.21\% to 53.37\% on \emph{mini}ImageNet.
\end{abstract}

\section{Introduction}
The recent advances in deep Convolutional Neural Networks (CNNs) have demonstrated high capability in visual recognition. For instance, an ensemble of residual nets \cite{he2016deep} achieves 3.57\% top-5 error on the ImageNet test set, which is even lower than 5.1\% of the reported human-level performance. The achievements have relied on the fact that learning deep CNNs requires large quantities of annotated data. As a result, the standard optimization of deep CNNs does not offer a satisfactory solution for learning new categories from very little data, which is generally referred to as ``One-Shot or Few-Shot Learning" problem. One possible way to alleviate this problem is to capitalize on the idea of transfer learning \cite{Bengio:ICML12,Yosinski:NIPS14} by fine-tuning a pre-trained network from another task with more labelled data. However, as pointed out in \cite{Yosinski:NIPS14}, the benefit of a pre-trained network will greatly decrease especially when the network was trained on the task or data which is very different from the target one, not to mention that the very little data may even break down the whole network due to overfitting. More importantly, the general training procedure which contains a number of examples per category in each batch does not match inference at test time when only a single or very few examples of a new category is given. This discrepancy affects the generalization of the learnt deep CNNs from prior knowledge.

We propose to mitigate the aforementioned two issues in our one-shot learning framework. First, we induce from a single or few examples per category to form a small set of labelled images (support set) in each batch of training. The optimization of our framework is then performed by recognizing other instances (unlabelled images) from the categories in the support set correctly. As such, the training strategy is amended particularly for one-shot learning so as to match inference in the test stage. Moreover, a memory module is leveraged to compress and generalize the input set into slots in the memory and produce the outputs holistically on the whole support set, which further enhances the recognition. Second, we feed the memory slots into one Recurrent Neural Networks (RNNs), as a contextual learner, to predict the parameters of CNNs for the unlabelled images. As a result, the contextual learner captures both long-term memory across all the categories in the training and short-term knowledge specified on the categories at test time. Note that our solution does not require a fine-tuning process and computes the parameters on the fly. In addition, the memory is an uniform medium which could convert different size of support sets into common memory slots, making it very flexible to train an unified model irrespective of the number of shots and categories.

\begin{figure*}[!tb]
	\centering {\includegraphics[width=0.92\textwidth]{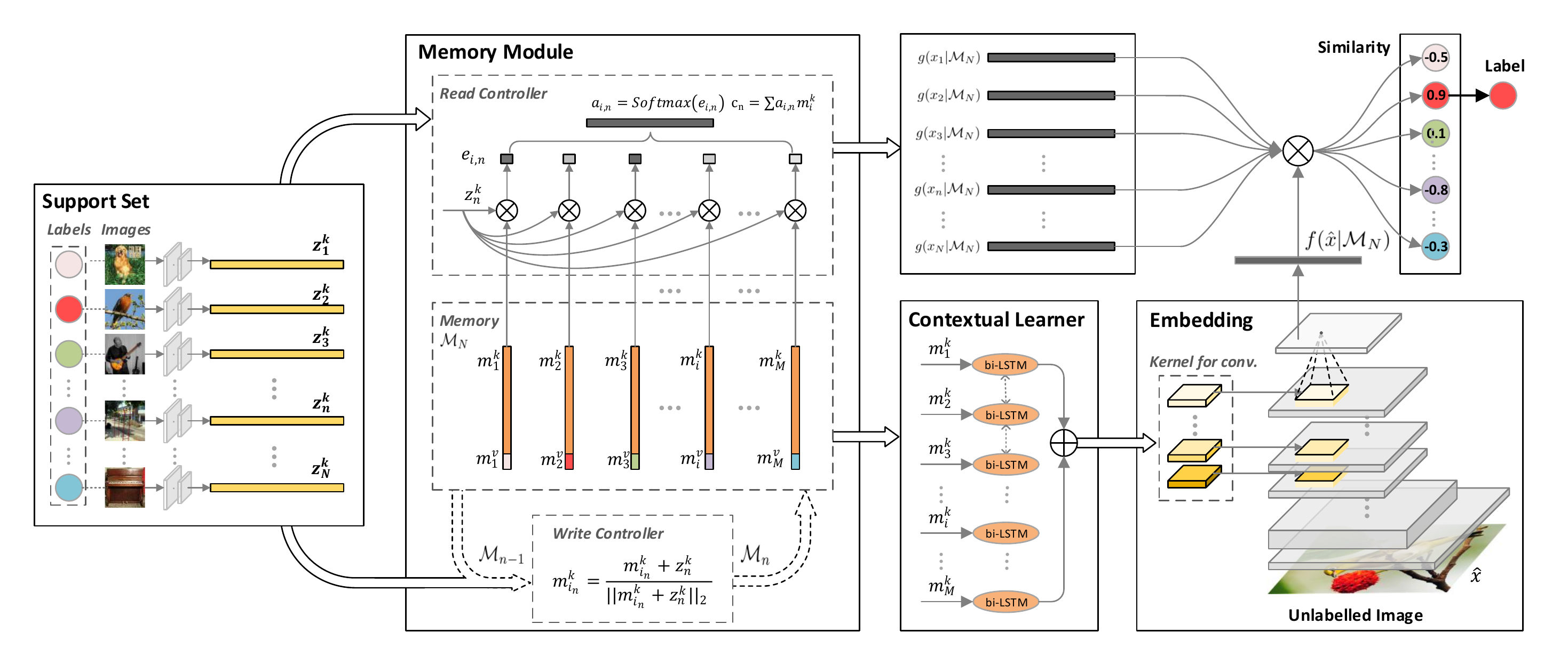}}
    \vspace{-0.125in}
	\caption{\small The overview of Memory Matching Networks (MM-Net) for one-shot image recognition (better viewed in color). Given a support set consisting of a single or few labelled examples per category, a deep CNNs is exploited to learn rich image representations, followed by a memory module to compress and generalize the input support set into slots in the memory via a write controller. A read controller in memory module further enhances the representation (embedding) learning of images in the support set by holistically exploiting the memory across all the categories. Meanwhile, a contextual learner, i.e., bi-LSTM, is adopted to explore the contextual relations between categories by encoding the memory slots in a sequential manner for predicting the parameters of CNNs, whose outputs are regarded as embeddings of unlabelled images. The dot product between the embeddings of a given unlabelled image and each image in the support set is computed as the similarity and the label of the nearest one is assigned to this unlabelled image. The training of our MM-Net exactly matches the inference. In addition, the memory is an uniform medium which could convert different size of support sets into common memory slots, making it flexible to train an unified model with a mixed strategy for performing inference on any one-shot learning scenarios.}
	\label{fig:fig1}
	\vspace{-0.2in}
\end{figure*}

By consolidating the idea of learning a learner to predict parameters in networks and matching training and inference strategy, we present a novel Memory Matching Networks (MM-Net) for one-shot image recognition, as shown in Figure \ref{fig:fig1}. Specifically, a single or few examples per category are fed into a batch every time as a support set of labelled images in training. A deep CNNs is exploited to learn image representations, which update the memory through a write controller. A read controller enhances the image representations with the memory across all the categories to produce feature embeddings of images in the support set. Meanwhile, we take the memory slots as a sequence of inputs to a contextual learner, i.e., bidirectional Long Short-Term Memory (bi-LSTM) networks, to predict the parameters of the convolutional layers in the CNNs. The outputs of CNNs are regarded as embeddings of unlabelled images. As such, the contextual relations between categories are also explored in learning network parameters. The dot product between the embeddings of a given unlabelled image and each image in the support set is computed as the similarity and the label of the nearest one is assigned to this unlabelled image. The whole deep network is end-to-end optimized by minimizing the error of predicting the labels in the batch conditioned on the support set. It is also worth noting that we could form each batch with different number of shots and categories in training stage to learn an unified architecture for performing inference on any one-shot learning scenarios. At inference time, the support set is then replaced by the examples from new categories and there is no any change in the procedure.

The main contribution of this work is the proposal of Memory Matching Networks for addressing the issue of one-shot learning in image recognition. The solution also leads to the elegant views of how the discrepancy between training and inference in one-shot learning should be amended and how to make the parameters of CNNs computable on the fly in the context of very little data, which are problems not yet fully understood in the literature.

\section{Related Work}\label{sec:RW}
\textbf{One-Shot Learning.}
The research of one-shot learning has proceeded mainly along following directions: data augmentation, transfer learning, deep embedding learning, and meta-learning. Data augmentation method \cite{dosovitskiy2014discriminative,hariharan2016low} is the most natural solution for one-shot learning by enlarging training data via data manufacturing. Transfer learning approaches \cite{fe2003bayesian, wang2016learning} aim to recycle the knowledge learned from previous tasks for one-shot learning. Wang \emph{et al.} exploit the generic category agnostic transformation from small-sample models to the underlying large-sample models for one-shot learning in \cite{wang2016learning}. Deep embedding learning \cite{koch2015siamese,vinyals2016matching} attempts to create a low-dimensional embedding space, where the transformed representations are more discriminative. \cite{koch2015siamese} learns the deep embedding space with a siamese network and classifies images by a nearest-neighbor rule. Later in \cite{vinyals2016matching}, Matching Network is developed to transform the support set and testing samples into a shared embedding space with matching mechanism. Meta-learning models \cite{bertinetto2016learning,ravi2017optimization,santoro2016meta} mainly frame the learning problem at two levels: the rapid learning to acquire the knowledge within each task and the gradual learning to extract knowledge learned across all tasks. For instance, \cite{ravi2017optimization} proposes an LSTM-based meta-learner model to learn the exact optimization algorithm, which is utilized to train another neural network classifier in the few-shot regime.

\textbf{Parameter Prediction in CNNs.}
Parameter prediction in CNNs refers to evolve one network to generate the structure of weights for another network. \cite{schmidhuber1992learning} is one of the early works that suggests the concept of fast weights in which one network can produce the changes of context-dependent weights for a second network. Later in \cite{denil2013predicting}, Denil \emph{et al.} demonstrate the significant redundancy in the parameterization of several deep learning models and it is possible to accurately predict most parameters given only a few weights. Next, a few subsequent works study practical applications with the fast weights concept, e.g., image question answering \cite{noh2016image} and zero-shot image recognition \cite{lei2015predicting}.

\textbf{Memory Networks.}
Memory Networks is first proposed in \cite{weston2014memory} by augmenting neural networks with an external memory component which can be easily read and written through read and write controllers. Later in \cite{sukhbaatar2015end}, Memory Networks is further extended to End-to-end Memory Networks, which is trained in an end-to-end manner and requires significantly less supervision compared with original Memory Networks. Moreover, Chandar \emph{et al.} explore a form of Hierarchical Memory Networks \cite{chandar2016hierarchical}, allowing the read controller to efficiently access extremely large memories. Recently, Key-Value Memory Networks \cite{miller2016key} stores prior knowledge in a key-value structured memory before reading them for prediction, making the knowledge to be stored more flexibly. In this work, we adopt the Key-Value Memory Networks as the memory module to store the encoded contextual information specified on the categories into the key-value structured memory.

In summary, our work belongs to deep embedding learning method for one-shot learning. However, most of the above methods in this direction mainly focus on forming the deep embedding space with the simple objective of matching-based classification (i.e., to maximize the matching score between unlabelled image and the support images with the same label). Our work is different that we enhance the one-shot learning by leveraging memory module to additionally integrate the contextual information across support samples into the deep embedding architectures. It is worth noting that \cite{vinyals2016matching} also involves contextual information for one-shot learning. Ours is fundamentally different in the way that all the CNNs in \cite{vinyals2016matching} need to be learnt at training stage, as opposed to directly predicting the parameters of CNNs for unlabelled image based on the contextual information encoded in the memory slots of this work, which is better-suited for one-shot learning during inference on unseen categories.

\section{One-Shot Image Recognition}
The basic idea of Memory Matching Networks (MM-Net) for one-shot learning is to construct an embedding space where the unseen objects can be rapidly recognized from a few labelled images (support set). MM-Net firstly utilizes a memory module to encode and generalize the whole support set into memory slots, which are endowed with the contextual information specified on the categories. The training of MM-Net is then performed by contextually embedding the whole support set with the memory across all the categories via read controller. Meanwhile, a contextual learner is devised to predict the parameters of CNNs for embedding unlabelled image conditioned on the contextual relations between categories. Both of the embeddings of support set and unlabelled image are further leveraged to retrieve the label of unlabelled image through matching mechanism in the embedding space. Our MM-Net is trained in a learning to learn manner and can be adapted flexibly for recognizing any new objects by only feed-forwarding the support set. An overview of MM-Net is shown in Figure \ref{fig:fig1}.

\subsection{Problem Formulation}
Suppose we have a small support set with $N$ image-label pairs $\mathcal{S} = \{(x_n,y_n)\}^N_{n=1}$ from $C$ object classes, where each class contains few or even one single image. In the standard setting of one-shot learning, our ultimate target is to recognize a class from a single labelled image. Hence, given an unlabelled image $\hat{x}$, we aim to predict its class $\hat{y}$ with the prior knowledge mined from the support set $\mathcal{S}$, which is defined as
\begin{equation}\label{Eq:EqPF1}\small
	\hat{y} = \mathop {\arg \max }\limits_{y_n \in {\mathcal{C}}} P\left( {y_n|\hat{x},\mathcal{S}} \right),
	\vspace{-0.05in}
\end{equation}
where $P\left( {y_n|\hat{x},\mathcal{S}} \right)$ is the probability of classifying $\hat{x}$ with the class $y_n$ conditioned on $\mathcal{S}$ and $\mathcal{C}$ is the set of class labels. Inspired by the recent success of Matching Networks in one-shot learning \cite{vinyals2016matching}, we formulate our one-shot object recognition model in a non-parametric manner based on matching mechanism which retrieves the class label of unlabelled image by comparing the matching scores with all the labelled images (support set) in the learnt embedding space. Accordingly, the probability of classifying $\hat{x}$ with the class label $y_n$ we exploit here can be interpreted as the matching score between $\hat{x}$ and the support sample $x_n$ with label $y_n$, which is measured as the dot product between their embedded representations
\begin{equation}\label{Eq:EqPF2}\small
	P\left( {{y_n}|\hat x,\mathcal{S}} \right) = f{\left( {\hat x|\mathcal{S}} \right)^\top} \cdot g\left( {{x_n}|\mathcal{S}} \right),
	\vspace{-0.05in}
\end{equation}
where $f\left(\cdot\right)$ and $g\left(\cdot\right)$ are two deep embedding functions for unlabelled image $\hat{x}$ and support image $x_n$ given the whole support set $\mathcal{S}$, respectively. Please note that derived from the idea of Memory Networks \cite{weston2014memory}, we leverage a memory module to explicitly generalize the whole support set into memory slots, which are endowed with the contextual information among support set $\mathcal{S}$ and can be further integrated into the learning of both $f\left(\cdot\right)$ and $g\left(\cdot\right)$.

\subsection{Encoding Support Set with Memory Module}
Inspired from the recent success of Recurrent Neural Networks (RNNs) for sentence modeling in machine translation \cite{Bahdanau14} and image/video captioning \cite{pan2016jointly,yao2017boosting}, one natural way to model the contextual relationship across the support samples in support set $\mathcal{S}$ is to adopt the RNNs based models as in \cite{ravi2017optimization,vinyals2016matching}, whose latent state is treated as the memory. However, such kind of memory is typically too small and not compartmentalized enough to accurately remember the previous knowledge, let alone the contextual information across diverse object classes with few or even one single image per class. Taking the inspiration from Memory Networks \cite{weston2014memory} which manipulates a large external memory that can be flexibly read and written to, we design a memory module to encode the contextual information within support set into the memory through write controller.

\textbf{Memory.}
The memory in our memory module is denoted as $\mathcal{M} = \{(m^k_i,m^v_i)\}^M_{i=1}$ consisting of $M$ key-value pairs, where each memory slot is composed of a memory key $m^k_i$ and the corresponding memory value $m^v_i$. Here the memory key $m^k_i\in {{\mathbb{R}}^{{{D}_m}}}$ denotes the ${D}_m$-dimensional memory representation of the $i$-th memory slot and the memory value $m^v_i$ is an integer representing the class label of the $i$-th memory slot.

\textbf{Write controller.}
Given the support set $\mathcal{S}$, the memory module is utilized to encode the sequence of $N$ support images into $M$ memory slots with write controller, aiming to distill the intrinsic characteristics of classes. Thus, we devise the memory updating strategy in our write controller as a dynamic feature aggregation problem to exploit both the intrinsic universal characteristic of each class beyond individual samples and the remarkable diversity within each class. The core issue for this design is about whether the write controller should jointly aggregate visually similar support samples into one memory slot by sequentially updating the corresponding memory key or individually seek one new memory slot to store the distinctive samples. The former one is triggered when the input support sample shares the same class label/memory value with the visually similar memory key, otherwise the later one is adopted.

The vector formulas for the memory updating strategy in write controller are given below. At $n$-th time step, the current input support image $x_n$ and its class label $y_n$ are written into memory slots to update the previous memory $\mathcal{M}_{n-1}$ via write controller, producing memory $\mathcal{M}_{n}$. In particular, let $z_n\in {{\mathbb{R}}^{{{D}_z}}}$ denote the $D_z$-dimensional visual feature of the support image $x_n$. One transformation matrices ${{\bf{T}}_z} \in {{\mathbb{R}}^{{{D}_m} \times {{D}_z}}}$ is firstly employed to project the support image $z_n$ into the mapping $z_{n}^{k}$ in memory key space:
\begin{equation}\label{Eq:EqUpdate}\small
	z_{n}^{k} ={{\bf{T}}_z} z_{n}.
\end{equation}
Next, for the input support image, we mine its nearest neighbor (i.e., the most visually similar memory key) from previous memory $\mathcal{M}_{n-1}$ with respect to dot product similarity between its representation in memory key space $z_{n}^{k}$ and each memory key $m^k_i$. Here we denote $i_n$ as the index of $x_n$'s nearest neighbor in memory $\mathcal{M}_{n-1}$. The memory updating is then conducted in a different way depending on whether the memory value of $x_n$'s nearest neighbor $m^v_{i_n}$ is exactly matched with the $x_n$'s class label $y_n$ or not. If $m^v_{i_n}=y_n$, we only update the memory key $m_{i_{n}}^{k}$ by integrating it with $z_{n}^{k}$ and then normalizing it:
\begin{equation}\label{Eq:EqUpdate2}\small
	m_{{i_n}}^k = {{\left( {m_{{i_n}}^k + z_n^k} \right)} \mathord{\left/
	{\vphantom {{\left( {m_{{i_n}}^k + z_n^k} \right)} {||m_{{i_n}}^k + z_n^k|{|_2}}}} \right.
	\kern-\nulldelimiterspace} {||m_{{i_n}}^k + z_n^k|{|_2}}}.
\end{equation}
Otherwise, when $m^v_{i_n} \neq y_n$, we store the key-value pair $(z_{n}^{k}, y_n)$ in the next new memory slot. Note that if there is no available memory slot left, the memory key $m_{i_{n}}^{k}$ is updated as in Eq.(\ref{Eq:EqUpdate2}). After encoding the whole support set $\mathcal{S}$ into memory via write controller, the final memory ${\mathcal{M}}_{N}$ is endowed with the contextual information within support set. Please note that we denote the two deep embedding functions $f{\left( {\hat x|\mathcal{S}}\right)}$ and $g{\left( {x_{n}}|\mathcal{S}\right)}$ as $f{\left( {\hat x|\mathcal{M}_{N}} \right)}$ and $g{\left( {x_{n}}|\mathcal{M}_{N}\right)}$ in the following sections, respectively.

\subsection{Contextual Embedding for Support Set}\label{sec:support}
The most typical way to transform images from the support set into the embedding space is to embed each sample independently through a shared deep embedding architecture $g{\left( {x_n} \right)}$ in discriminative learning, while the holistical contextual information within support set is not fully exploited. Here we develop a contextual embedding function for support set $g\left( {{x_n}|\mathcal{M}_{N}}\right)$ to embed $x_n$ conditioned on the memory $\mathcal{M}_{N}$ via read controller of memory module, with the intuition that the holistical contextual information endowed in the memory across all the categories can guide $g$ to produce more discriminative representation of $x_n$.

\textbf{Read controller.} Technically, for each support image $x_n$ and its embedded representation ${z}^{k}_{n}$ in memory key space, we firstly measure the dot product similarity $a_{i,n}$ between ${z}^{k}_{n}$ and each memory key ${m}^{k}_{i}$ followed by a softmax function, and then retrieve the aggregated memory vector $c_n$ by calculating the sum of each memory key weighted by $a_{i,n}$:
\begin{equation}\label{Eq:CE1}\small
	a_{i,n} = \verb|Softmax| ({{z}^{k}_{n}}^\top \cdot {m}^{k}_{i}),~~~ {c_n} = \sum\nolimits_{i = 1}^M {{a_{i,n}} m_i^k},
\end{equation}
where $\verb|Softmax|(v_i)={{{e^{{v_i}}}}/{\sum\nolimits_j {{e^{{v_j}}}} }}$. The above memory retrieval process is conducted by read controller. Besides, a shortcut connection is additionally constructed between the input and output of read controller, making the optimization more easier. Thus, the final output representation of $x_{n}$ via contextual embedding is measured as:
\begin{equation}\label{Eq:CE2}\small
	{g(x_n|\mathcal{M}_{N})}= {{\bf{T}}_c} c_n + z_{n} \in {{\mathbb{R}}^{D_z}},
\end{equation}
where ${{{\bf{T}}_c}} \in {{\mathbb{R}}^{{{D}_z} \times {{D}_m}}}$ is the transformation matrix for mapping the aggregated memory $c_n$ into the embedding space and $D_z$ is the embedding space dimension.

\subsection{Contextual Embedding for Unlabelled Images}\label{sec:learner}
The standard deep embedding function $f{\left( {\hat x;W} \right)}$ in discriminative learning consists of stacks of convolutional layers that are parameterized by matrix $W$ in general. The optimization of parameters $W$ often requires enormous training data and a lengthy iterative process to generalize well on unseen samples. However, in the extreme case with only a single labelled example of each class, it is insufficient to train the deep embedding architecture and directly fine-tuning this architecture often results in poor performance on the recognition of new category. To address the aforementioned challenges for one-shot learning, we devise a novel contextual embedding architecture $f{\left( {\hat x;W|\mathcal{M}_{N}} \right)}$ for unlabelled image by incorporating the contextual relations between categories mined from memory $\mathcal{M}_{N}$ into the deep embedding function. In particular, the parameters $W$ of this contextual embedding architecture are learnt in a feed-forward manner conditioned on memory $\mathcal{M}_{N}$ without backpropagation, obviating the need of fine-tuning to adapt to the new category.

\textbf{Contextual Learner.} A novel deep architecture, named as contextual learner, is especially designed to synthesis the parameters $W$ of contextual embedding architecture $f{\left( {\hat x;W|\mathcal{M}_{N}} \right)}$ depending on the memory $\mathcal{M}_{N}$ of support set. Specifically, we denote the output parameters $W\in {\mathbb{R}}^{D_w}$ for contextual learner as
\begin{equation}\label{Eq:EqCES1}\small
	W = \omega (\mathcal{M}_{N};W'),
\end{equation}
where $\omega\left(\cdot\right)$ is the encoding function of contextual learner that transforms the memory $\mathcal{M}_{N}$ into the target parameters $W$ and $W'$ are the parameters of contextual learner $\omega(\cdot)$. Inspired by the success of bidirectional LSTM (bi-LSTM) \cite{schuster1997bidirectional} in several  inherently sequential tasks (e.g., machine translation \cite{Bahdanau14}, speech recognition \cite{bahdanau2016end,graves2013hybrid} and video generation \cite{pan2017to}), we leverage bi-LSTM to contextually encode the memory $\mathcal{M}_{N}$ in a sequential manner. In particular , bi-LSTM consisting of forward and backward LSTMs \cite{Hochreiter:NC97}, which read the memory slots of $\mathcal{M}_{N}$ in its natural order (from $m_{1}^{k}$ to $m_{M}^{k}$) and the reverse order (from $m_{M}^{k}$ to $m_{1}^{k}$), respectively. The encoded representation ${\bf{\overleftrightarrow{h}}}\in {{\mathbb{R}}^{D_r}}$ for memory $\mathcal{M}_{N}$ is achieved by directly summating the final hidden states of two LSTMs, where $D_r$ denotes the dimension of LSTM hidden state. The output parameters $W$ are calculated as
\begin{equation}\label{Eq:EqCES3}\small
	W = {{\bf{T}}_p}{\bf{\overleftrightarrow{h}}},
\end{equation}
where ${{\bf{T}}_p} \in {{\mathbb{R}}^{{{D}_w} \times {{D}_r}}}$ is the transformation matrix. Accordingly, by synthesizing the parameters of contextual embedding with our contextual learner, the contextual relations between categories are elegantly integrated into this deep embedding architecture $f{\left( {\hat x;W|\mathcal{M}_{N}} \right)}$ for the unlabelled image, which encourages the transformed representation to be more discriminative for image recognition.

\textbf{Factorized Architectures.}
When designing the specific architecture of the contextual embedding for unlabelled images, the traditional convolutional layer is modified with factorized design \cite{bertinetto2016learning} for significantly reducing the number of parameters within convolutional filters, making parameter prediction with contextual learner more feasible.

\subsection{Training Procedure}\label{training procedure}
After obtaining the embedded representations of both unlabelled image and the whole support set, we follow the prior works \cite{ravi2017optimization,vinyals2016matching} to train our model for the widely adopted task of one-shot learning: the $C$-way $k$-shot image recognition task, i.e., classifying a disjoint set of unlabelled images given a set of $C$ unseen classes with only $k$ labelled images per class. Specifically, for each batch in the training stage, we firstly sample $C$ categories uniformly from all training categories with $k$ examples per category, forming the labelled support set $\mathcal{S}$. The corresponding unlabelled images set $\mathcal{B}$ are randomly sampled from the rest data belonging to the $C$ categories in training set. Hence, given the support set $\mathcal{S}$ and input unlabelled images set $\mathcal{B}$, the softmax loss is then formulated as:
\begin{equation}\label{Eq:TP1}\scriptsize
	{\mathcal{L}}(\mathcal{S}, \mathcal{B}) =  - \sum\limits_{{\hat x} \in \mathcal{B}} {\sum\limits_{\left( {{x_n},{y_n}} \right) \in \mathcal{S}} {{\mathbb{I}_{\left( {{\hat y} = {y_n}} \right)}} \log \frac{{{e^{P\left( {{y_n}|\hat x,\mathcal{S}} \right)}}}}{{\sum\limits_{\left( {{x_t},{y_t}} \right) \in \mathcal{S}} {{e^{P\left( {{y_t}|\hat x,\mathcal{S}} \right)}}} }}}},
\end{equation}
where $\hat y \in \mathcal{C}$ represents the class label of $\hat{{x}}$ and $P\left( {{y_n}|\hat x,\mathcal{S}} \right)$ denotes the probability of classifying $\hat{{x}}$ with the class label of $x_{n}$ as in Eq.(\ref{Eq:EqPF2}). The indicator function $\mathbb{I}_\emph{condition}=1$ if $\emph{condition}$ is true; otherwise $\mathbb{I}_\emph{condition}=0$. By minimizing the softmax loss over a training batch, our MM-Net is trained to recognize the correct class labels of all the images in $\mathcal{B}$ conditioned on the support set $\mathcal{S}$. Accordingly, in the test stage, given the support set $\mathcal{S}'$ containing $C$ categories never seen during training, our model can rapidly predict the class label for an unlabelled image through matching mechanism, without any fine-tuning on the novel categories due to its non-parametric property.

\textbf{Mixed training strategy.} In the above mentioned training procedure, each training batch is constructed with the uniform setting which exactly matches the test setting ($C$-way $k$-shot), targeting for mimicking the test situation for one-shot learning. However, such matching mechanism indicates that the learnt model is only suitable for the pre-fixed $C$-way $k$-shot test scenario, making it difficult to generalize to other $C'$-way $k'$-shot task (where $C' \not= C$ or $k' \not= k$). Accordingly, to enhance the generalization of the one-shot learning model, we devise a mixed training strategy by constructing each training batch with different number of shots and categories to learn an unified architecture for performing inference on any one-shot learning scenarios. Please note that the memory could be regarded as an uniform medium which converts different size of support sets into common memory slots. As a result, the mixed training strategy can be applied to learn an unified model irrespective of the number of shots and categories.

\section{Experiment}
We evaluate and compare our MM-Net with state-of-the-art approaches on two datasets, i.e., Omniglot \cite{lake2015human} and \emph{mini}ImageNet \cite{ravi2017optimization}. The former is the most popular one-shot image recognition benchmark of handwritten characters and the latter is a recently released subset of ImageNet \cite{russakovsky2015imagenet}.

\subsection{Datasets}
\textbf{Omniglot.} Omniglot contains 32,460 images of handwritten characters. It consists of 1,623 different characters within 50 alphabets ranging from well-established international languages like Latin and Korean to lesser-known local dialects. Each character was hand drawn by 20 different people via Amazon's Mechanical Turk, leading to 20 images per character. We follow the most common split in \cite{vinyals2016matching}, taking 1,200 characters for training and the rest 423 for testing. Moreover, the same data preprocessing in \cite{vinyals2016matching} is adopted, i.e., each image is resized to $28 \times 28$ pixels and rotated by multiples of $90$ degrees as data augmentation.

\textbf{\emph{mini}ImageNet.} The \emph{mini}ImageNet dataset is a recent collection of ImageNet for one-shot image recognition. It is composed of 100 classes randomly selected from ImageNet \cite{russakovsky2015imagenet} and each class contains 600 images with the size of $84\times 84$ pixels. Following the widely used setting in prior work \cite{ravi2017optimization}, we take 64 classes for training, 16 for validation and 20 for testing, respectively.

\subsection{Experimental Settings}
\textbf{Evaluation Metrics.}
All of our experiments revolve around the same basic task: the $C$-way $k$-shot image recognition task. In the test stage, we randomly select a support set consisting of $C$ novel classes with $k$ labelled images per class from the test categories and then measure the classification accuracy of the disjoint unlabelled images (15 images per class) for evaluation. To make the evaluation more convincing, we repeat such evaluation procedures 500 times for each setting and report the final mean accuracy for each setting. Moreover, the 95\% Confidence Intervals (CIs) of the mean accuracy is also present, which statistically describes the uncertainty inherent in performance estimation like standard deviation. The smaller the confidence interval, the more precise the mean accuracy performance.

\textbf{Network Architectures and Parameter Settings.}
For fair comparison with other baselines, we adopt a widely adopted CNNs in \cite{ravi2017optimization,vinyals2016matching} as the embedding function for support set $g\left(\cdot\right)$, consisting of four convolutional layers. Each convolutional layer is devised with a $3 \times 3$ convolution with 64 filters followed by batch normalization, a ReLU non-linearity and a $2 \times 2$ max-pooling. Accordingly, the final output embedding space dimension $D_z$ is 64 on Omniglot and 1,600 on \emph{mini}ImageNet, respectively. The contextual embedding for unlabelled image $f\left(\cdot\right)$ is similar to $g\left(\cdot\right)$ except that the last convolution layer is developed with factorized design and its parameters are predicted based on the contextual memory of support set. For the memory module, the dimension of each memory key $D_m$ is set as 512. For contextual learner, we set the size of hidden layer in bi-LSTM as 512. Our MM-Net is trained by Adam \cite{kingma2014adam} optimizer. The initial learning rate is set as 0.001 and we decrease it to 50\% every 20,000 iterations. The batch size is set as 16 and 4 for Omniglot and \emph{mini}ImageNet.

\subsection{Compared Approaches}
To empirically verify the merit of our MM-Net model, we compare with the following state-of-the-art methods: (1) Siamese Networks (SN) \cite{koch2015siamese} optimizes siamese networks with weighted $L_1$ loss of distinct input pairs for one-shot learning. (2) Matching Networks (MN) \cite{vinyals2016matching}  performs one-shot learning with matching mechanism in the embedding space, which is further developed into fully-contextual embedding version (MN-FCE) by utilizing bi-LSTM to contextually embed samples. (3) Memory-Augmented Neural Networks (MANN) \cite{santoro2016meta} devises a memory-augmented neural network to rapidly assimilate new data for one-shot learning. (4) Model-Agnostic Meta-Learning (MAML) \cite{finn2017model} learns easily adaptable model parameters through gradient descent in a meta-learning fashion. (5) Meta-Learner LSTM (ML-LSTM) \cite{ravi2017optimization} designs a LSTM-based meta-learner to learn an update rule for optimizing the network. (6) Siamese with Memory (SM) \cite{kaiser2017learning} presents a life-long memory module to remember past training samples and makes predictions based on stored previous samples. (7) Meta-Networks (Meta-N) \cite{munkhdalai2017meta} takes the loss gradient as meta information to rapidly generate the parameters of classification networks. (8) Memory Matching Networks (MM-Net) is the proposal in this paper. Moreover, a slightly different version of this run is named as MM-Net$^-$, which is trained without the mixed training strategy.

\subsection{Results on Omniglot}
Table \ref{tab:res_omniglot} shows the performances of different models on Omniglot dataset. Overall, the results across 1-shot and 5-shot learning on 5 and 20 categories consistently indicate that our proposed MM-Net achieves superior performances against other state-of-the-art techniques including deep embedding models (SN, MN, SM) and meta-learning approaches (MANN, Meta-N, MAML). In particular, the 5-way and 20-way accuracy of our MM-Net can achieve 99.28\% and 97.16\% on 1-shot learning, making the absolute improvement over the best competitor Meta-N by 0.33\% and 0.16\%, respectively, which is generally considered as a significant progress on this dataset. As expected, the 5-way and 20-way accuracies are boosted up to 99.77\% and 98.93\% respectively when provided 5 labelled images (5 shot) from each category. SN, which simply achieves the deep embedding space through pairwise learning, is still effective in 5-way task. However, the accuracy is decreased sharply when searching nearest neighbor in the embedding space in 20-way 1-shot scenario. Furthermore, MN, MANN, SM, Meta-N, MAML, and MM-Net lead to a large performance boost over SN, whose training strategy does not match the inference. The results basically indicate the advantage of bridging the discrepancy between how the model is trained and exploited at test time. SM by augmenting CNNs with a life-long memory module to exploit the contextual memory among previous labelled samples for one-shot learning, improves MN, but the performances are still lower than our MM-Net. This confirms the effectiveness of the contextual learner for directly synthesizing the parameters of CNNs, obviating adapting the embedding to novel classes with fine-tuning.
\setlength\tabcolsep{1pt}
\begin{table}[!tb]\footnotesize
	\centering
	\begin{center}
		\caption{Mean accuracy (\%) $\pm$ CIs (\%) of our MM-Net and other state-of-the-art methods on Omniglot dataset.}
\vspace{-0.1in}
		\label{tab:res_omniglot}
		\begin{tabular}{l@{\hskip\colspaceS}c@{\hskip\colspaceS}c@{\hskip\colspaceS}c@{\hskip\colspaceS}c}
			\toprule
			\multirow{2}{*}{\textbf{Model}} & \multicolumn{2}{c}{\textbf{5-way Accuracy} } & \multicolumn{2}{c}{\textbf{20-way Accuracy} } \\ \cline{2-5}
											 & 1-shot                    & 5-shot                     & 1-shot                    & 5-shot                    \\
			\midrule

			SN \cite{koch2015siamese}        & $97.3$                    & $98.4$                     & $88.2$                    & $97.0$                    \\

			MN \cite{vinyals2016matching}    & $98.1$                    & $98.9$                     & $93.8$                    & $98.5$                    \\

			MANN \cite{santoro2016meta}      & $82.8$                    & $94.9$                     & ---                       & ---                       \\

			SM \cite{kaiser2017learning}     & $98.4$                    & $99.6$                     & $95.0$                    & $98.6$                    \\

			Meta-N \cite{munkhdalai2017meta} & $98.95$                   & ---                        & $97.00$                   & ---                       \\

			MAML \cite{finn2017model}        & $98.7 \pm 0.4 $           & \textbf{ 99.9 $\pm$ 0.1  } & $95.8 \pm 0.3  $          & \textbf{98.9 $\pm$ 0.2  } \\
			\midrule
			MM-Net                              & \textbf{99.28 $\pm$ 0.08} & 99.77 $\pm$ 0.04  & \textbf{97.16 $\pm$ 0.10} & \textbf{98.93 $\pm$ 0.05} \\
			\bottomrule
		\end{tabular}
	\end{center}
\vspace{-0.25in}
\end{table}

\subsection{Results on \emph{mini}ImageNet}
The performance comparisons on \emph{mini}ImageNet are summarized in Table \ref{tab:miniImagenet}. Our MM-Net performs consistently better than other baselines. In particular, the 5-way accuracies of 1-shot and 5-shot learning can reach 53.37\% and 66.97\%, respectively, which is to-date the highest performance reported on \emph{mini}ImageNet, making the absolute improvement over MAML by 4.67\% and 3.86\%. MN-FCE exhibits better performance than MN, by further taking contextual information within support set into account for embedding learning of images. ML-LSTM and MAML which learns an update rule to fine-tune the CNNs or the easily adaptable parameters of CNNs could be generally considered as extensions of MN in a meta-learning fashion, resulting in better performance. There is a performance gap between Meta-N and our MM-Net$^-$. Though both runs involve the parameters prediction of CNNs, they are fundamental different in the way of parameters prediction. Meta-N predicts the parameters of the classification networks for unlabelled images based on the loss gradient of support set, while our MM-Net$^-$ leverages contextual information in memory to jointly predict the parameters of CNNs for unlabelled images and contextually encode support images. As indicated by our results, MM-Net$^-$ is benefited from the memory-augmented CNNs for both support set and unlabelled images, and leads to apparent improvements. In addition, MM-Net by additionally leveraging the mixed training strategy outperforms MM-Net$^-$.

\setlength\tabcolsep{12pt}
\begin{table}[!tb]\footnotesize
	\centering
	\begin{center}
		\caption{Mean accuracy (\%) $\pm$ CIs (\%) of our MM-Net and other state-of-the-art methods on \emph{mini}ImageNet dataset.}
		\label{tab:miniImagenet}
		\begin{tabular}{l@{\hskip\colspaceDD}c@{\hskip\colspaceDD}c}
			\toprule
			\multirow{2}{*}{\textbf{Model}}  & \multicolumn{2}{c}{\textbf{5-way Accuracy}} \\ \cline{2-3}
												& 1-shot                    & 5-shot                    \\
			\midrule
			MN  \cite{vinyals2016matching}      & $43.40 \pm 0.78  $        & $51.09 \pm 0.71  $        \\
			MN-FCE  \cite{vinyals2016matching}  & $43.56 \pm 0.84  $        & $55.31 \pm 0.73  $        \\
			ML-LSTM \cite{ravi2017optimization} & $43.44 \pm 0.77  $        & $60.60 \pm 0.71  $        \\
			MAML \cite{finn2017model}           & $48.70 \pm 1.84  $        & $63.11 \pm 0.92  $        \\
			Meta-N \cite{munkhdalai2017meta}    & $49.21\pm0.96$            & ---                       \\
			\midrule
			MM-Net$^-$                          & $52.74\pm0.45 $           & $65.82\pm 0.37 $          \\
			MM-Net                                 & \textbf{53.37 $\pm$ 0.48} & \textbf{66.97 $\pm$ 0.35} \\
			\bottomrule
		\end{tabular}
	\end{center}
\vspace{-0.2in}
\end{table}

\setlength\tabcolsep{1pt}
\begin{table}[!tb]\footnotesize
	\begin{center}
		\caption{
			\label{tab:training_strategy}
			\small Mean accuracy (\%) of MM-Net by varying training strategies for $5$-way $k$-shot image recognition task ($k \in \left\{ {1,2,3,4,5} \right\}$) on \emph{mini}ImageNet.
		}
		\begin{tabular}{l@{\hskip\colspaceM}c@{\hskip\colspaceM}c@{\hskip\colspaceM}c@{\hskip\colspaceM}c@{\hskip\colspaceM}c@{\hskip\colspaceM}c@{\hskip\colspaceM}c@{\hskip\colspaceM}c}
			\toprule
			\multirow{2}{*}{\textbf{Train}} & \multicolumn{5}{c}{\textbf{Test}} \\ \cline{2-6}
										& 1-shot         & 2-shot          & 3-shot         & 4-shot          & 5-shot          \\
			\midrule
			{1-shot}                 & \textbf{52.74} & 57.53           & 59.31          & 60.02           & 60.33           \\
			{2-shot}                 & 52.68          & \textbf{59.14}  & 62.11          & 63.39           & 63.92           \\
			{3-shot}                 & 51.67          & 58.48           & \textbf{62.21} & 64.03           & 65.40           \\
			{4-shot}                 & 51.44          & 58.56           & 62.12          & \textbf{64.48}  & 65.77           \\
			{5-shot}                 & 51.09          & 58.03           & 61.80          & 64.14           & \textbf{65.82}  \\
			\midrule
			{Mixed $k$-shot}         & 52.83          & 59.88           & 63.31          & 65.32           & 66.71           \\
			{Mixed $C$-way $k$-shot} & \textbf{53.37} & \textbf{59.93 } & \textbf{63.35} & \textbf{65.49} & \textbf{66.97 } \\
			\bottomrule
		\end{tabular}
	\end{center}
\vspace{-0.25in}
\end{table}

\subsection{Experimental Analysis}
We further analyze the effect of training strategy, the hidden state size of bi-LSTM in contextual learner, the image representation embedding visualization, and the similarity matrix over test images for $5$-way $k$-shot image recognition task on \emph{mini}ImageNet Dataset.

\textbf{Training strategy.}
We first present the analysis to demonstrate the generalization of our MM-Net by employing mixed training strategy for various test scenarios. Table \ref{tab:training_strategy} details the performance comparisons between several training strategies (i.e., uniform and mixed training strategies) with respect to different test tasks (i.e., 1, 2, 3, 4 and 5-shot). Overall, for each test scenario, there is a clear performance gap between all the five uniform training strategies (i.e., 1, 2, 3, 4 and 5-shot) and our proposed mixed training strategies (i.e., Mixed $k$-shot and Mixed $C$-way $k$-shot). In particular, the peak performance of MM-Net is achieved when we adopt the Mixed $C$-way $k$-shot setting by changing both $C$ and $k$ ($C \in \left\{ {2,3,4,5} \right\}$, $k \in \left\{ {1,2,3,4,5} \right\}$) for constructing training batches. This empirically demonstrates the effectiveness of mixed training strategy for generalizing our MM-Net model to various test scenarios, obviating re-training the model on the new testing task. Note that Mixed $k$-shot, a simplified version of Mixed $C$-way $k$-shot which constructs each training batch with different number of shots but always in 5-way manner, still outperforms all the five uniform training strategies.

\begin{figure}[!tb]
	\centering {\includegraphics[width=0.41\textwidth]{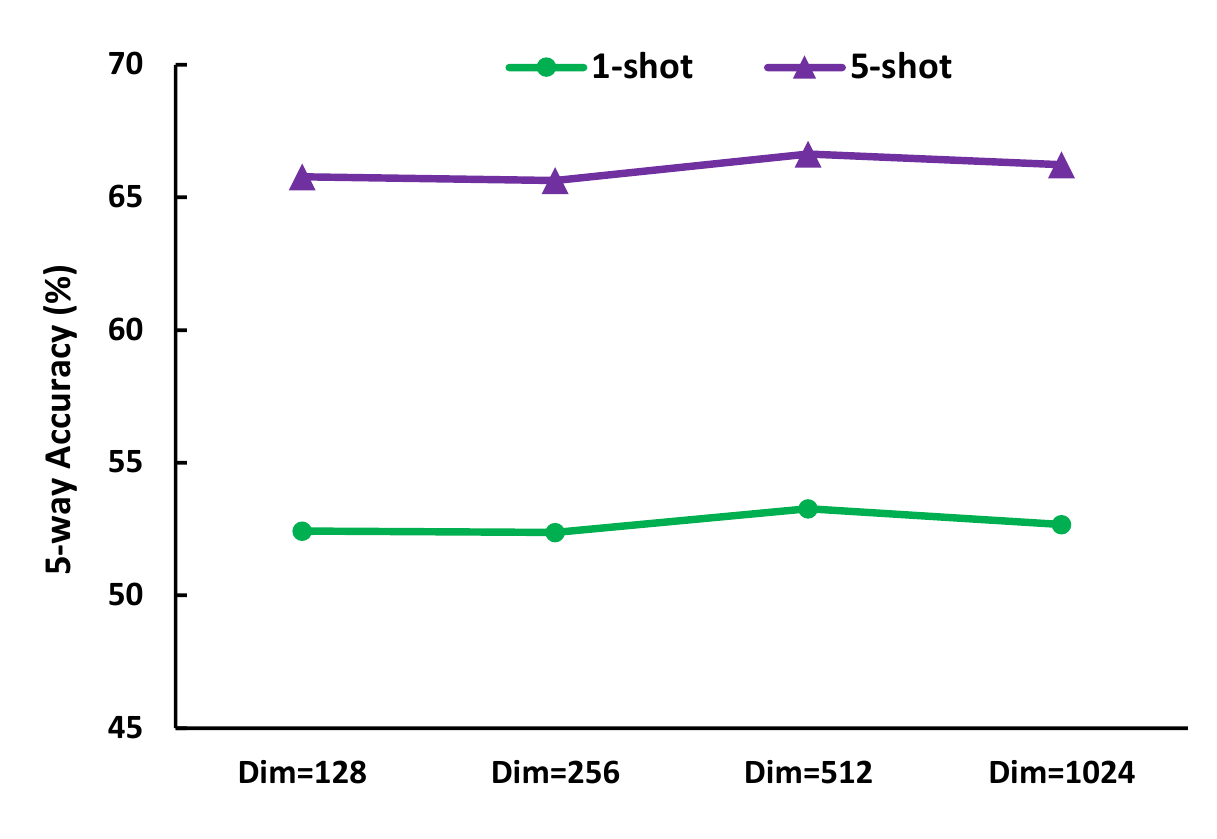}}
\vspace{-0.1in}
	\caption{\small The effect of the hidden state size in our contextual learner's bi-LSTM on \emph{mini}Imagenet.}
	\label{fig:lstm_size}
\vspace{-0.2in}
\end{figure}

\textbf{Hidden state size of bi-LSTM in contextual learner.}
In order to show the relationship between the performance and hidden state size of bi-LSTM in contextual learner, we compare the results of the hidden state size in the range of 128, 256, 512 and 1,024 on both 1-shot and 5-shot tasks. The 5-way accuracy with the different hidden state size is shown in Figure \ref{fig:lstm_size}. As illustrated in the figure, the performance difference by using different hidden state size is within 0.013 on both 1-shot and 5-shot tasks, which practically eases the selection for the optimal hidden state size.

\begin{figure}
	\centering
	\subfigure[MN \cite{vinyals2016matching}]{
		\label{fig:tsne:Matching-Networks}
		\includegraphics[width=0.23\textwidth]{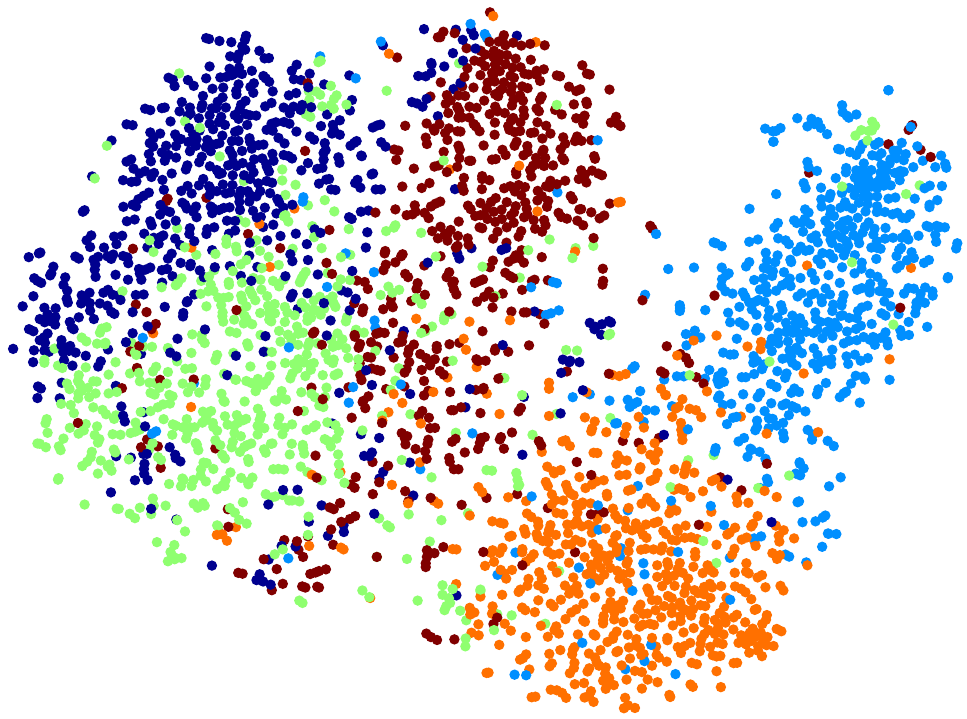}}
	\subfigure[MM-Net]{
		\label{fig:tsne:1-shot}
		\includegraphics[width=0.23\textwidth]{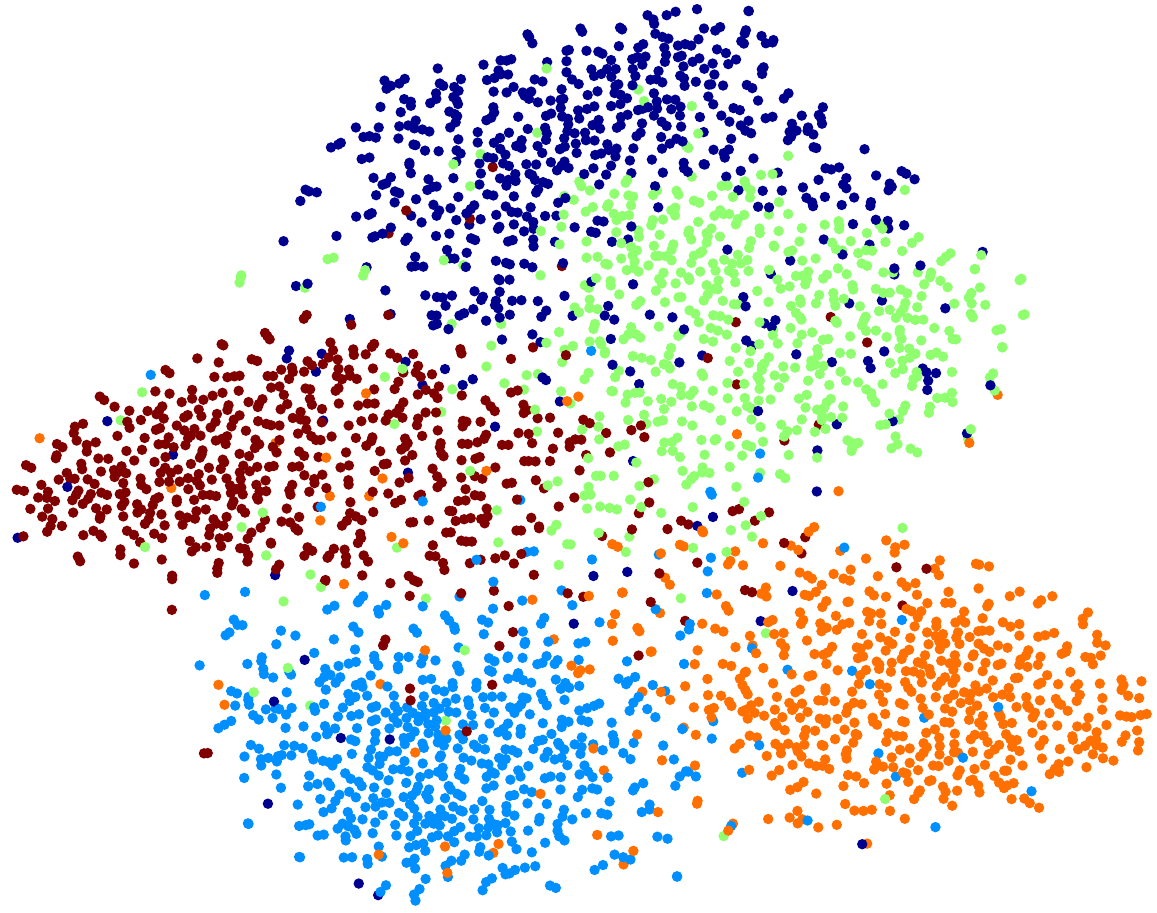}}
\vspace{-0.1in}	
\caption{\small Image representation embedding visualizations of MN and our MM-Net on \emph{mini}Imagenet using t-SNE \cite{maaten2008visualizing}. Each image is visualized as one point and colors denote different classes.}
	\label{fig:tsne}
\vspace{-0.2in}
\end{figure}

\textbf{Image representation embedding visualization.}
Figure \ref{fig:tsne} shows the t-SNE \cite{maaten2008visualizing} visualizations of embedding of image representation learnt by MN and our MM-Net under 5-way 5-shot scenario. Specifically, we randomly select 5 classes from \emph{mini}ImageNet testing set and the embedded representations of all the 2,975 images (excluding the 25 images in support set) are then projected into 2-dimensional space using t-SNE. It is clear that the embedded image representations by MM-Net are better semantically separated than those of MN.

\begin{figure}
	\centering
	\subfigure[MN \cite{vinyals2016matching}]{
		\label{fig:confusion:a}
		\includegraphics[width=0.22\textwidth]{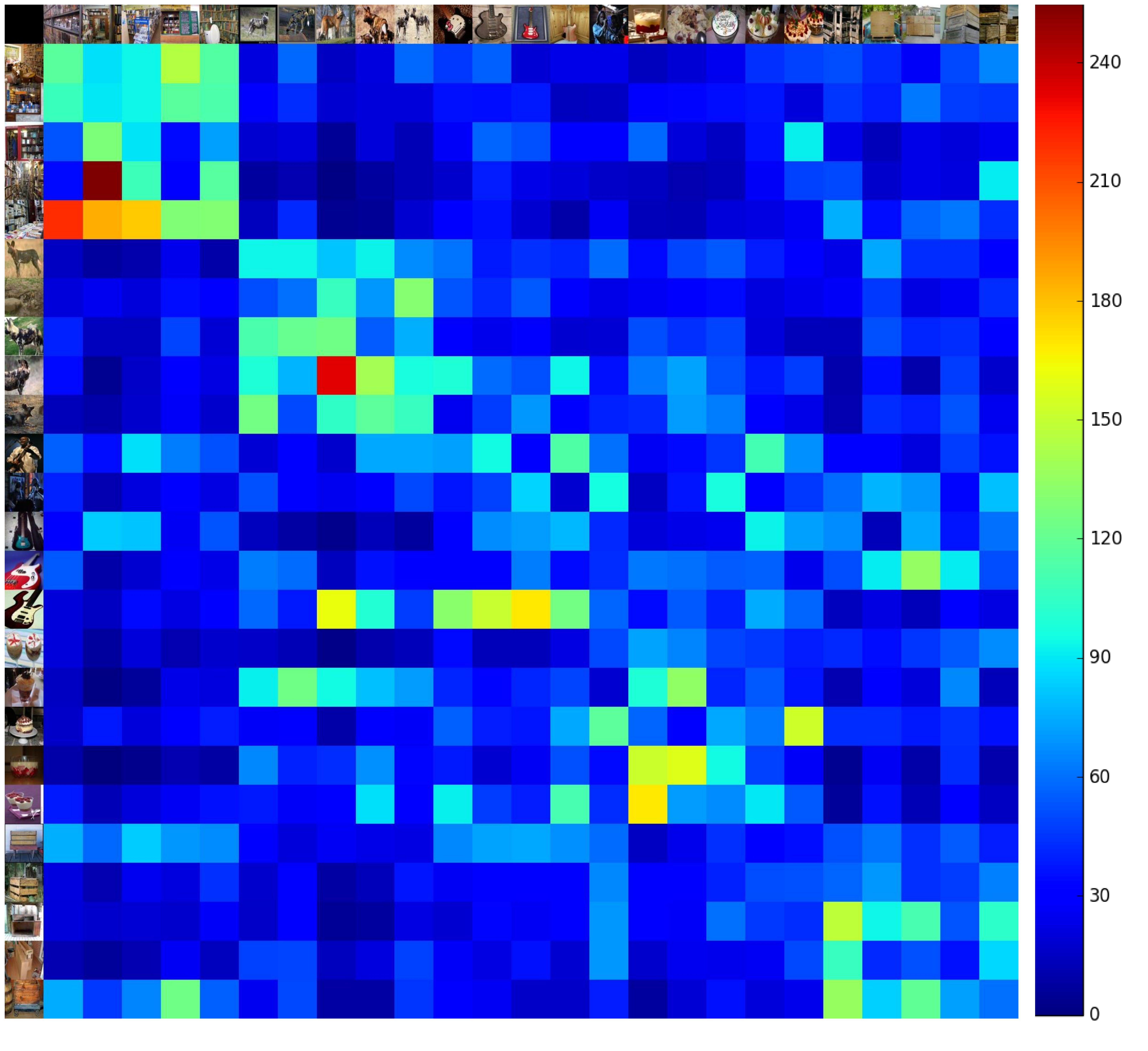}}
	\hspace{2mm}
	\subfigure[MM-Net]{
		\label{fig:confusion:b}
		\includegraphics[width=0.22\textwidth]{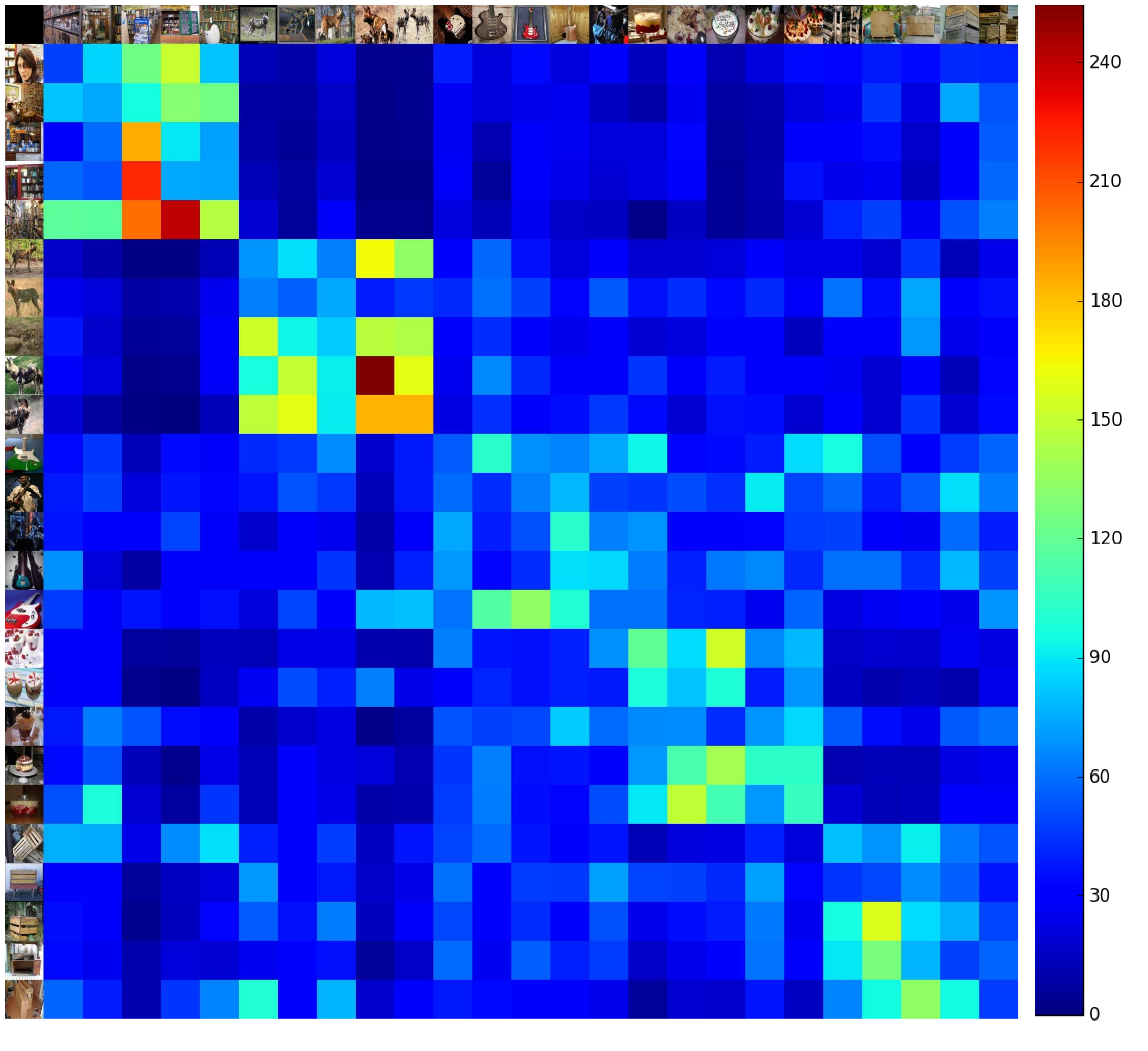}}
	\caption{\small Similarity matrix of MN and our MM-Net on \emph{mini}Imagenet (vertical axis: 5 labelled images per class in support set; horizontal axis: 5 unlabelled test images per class). The warmer colors indicate higher similarities.}
	\label{fig:confusion}
\vspace{-0.2in}
\end{figure}

\textbf{Similarity matrix visualization.}
Figure \ref{fig:confusion} further shows the visualizations of similarity matrix learnt by MN and our MM-Net under 5-way 5-shot scenario. In particular, the similarity matrix is constructed by measuring the dot product similarities between the randomly selected support set (25 images in 5 classes) and the corresponding 25 unlabelled test images. Note that every five images belong to the same class. Thus we can clearly see that most intra-class similarities of MM-Net are higher than those of MN and the inter-class similarities of MM-Net are mostly lower than MN, demonstrating that the representation learnt by our MM-Net are more discriminative for image recognition.

\section{Conclusions}
We have presented Memory Matching Networks (MM-Net), which explores a principled way of training the network to do one-shot learning as at inference. Particularly, we formulate the training by only utilizing one single or very few examples per category to form a support set of labelled images in each batch and switching the training from batch to batch, which is much like how it will be tested when presented with a few examples of new categories. Furthermore, through a new design of Memory module, the feature embeddings of images in the support set are contextually augmented with the holistic knowledge across categories in the set. Meanwhile, to better generalize the networks to the new categories with very little data, we construct a contextual learner which sequentially exploits the memory slots to predict the parameters of CNNs on the fly for unlabeled images. Experiments conducted on both Omniglot and \emph{mini}ImageNet datasets validate our proposal and analysis. Performance improvements are clearly observed when comparing to other one-shot learning techniques.

{\small
\bibliographystyle{ieee}
\bibliography{egbib}

\begin{thebibliography}{10}\itemsep=-1pt

\bibitem{Bahdanau14}
D.~Bahdanau, K.~Cho, and Y.~Bengio.
\newblock Neural machine translation by jointly learning to align and
  translate.
\newblock In {\em ICLR}, 2015.

\bibitem{bahdanau2016end}
D.~Bahdanau, J.~Chorowski, D.~Serdyuk, P.~Brakel, and Y.~Bengio.
\newblock End-to-end attention-based large vocabulary speech recognition.
\newblock In {\em ICASSP}, 2016.

\bibitem{Bengio:ICML12}
Y.~Bengio.
\newblock Deep learning of representations for unsupervised and transfer
  learning.
\newblock In {\em ICML Workshop on Unsupervised and Transfer Learning,}, 2012.

\bibitem{bertinetto2016learning}
L.~Bertinetto, J.~F. Henriques, J.~Valmadre, P.~Torr, and A.~Vedaldi.
\newblock Learning feed-forward one-shot learners.
\newblock In {\em NIPS}, 2016.

\bibitem{chandar2016hierarchical}
S.~Chandar, S.~Ahn, H.~Larochelle, P.~Vincent, G.~Tesauro, and Y.~Bengio.
\newblock Hierarchical memory networks.
\newblock {\em arXiv preprint arXiv:1605.07427}, 2016.

\bibitem{denil2013predicting}
M.~Denil, B.~Shakibi, L.~Dinh, N.~de~Freitas, et~al.
\newblock Predicting parameters in deep learning.
\newblock In {\em NIPS}, 2013.

\bibitem{dosovitskiy2014discriminative}
A.~Dosovitskiy, J.~T. Springenberg, M.~Riedmiller, and T.~Brox.
\newblock Discriminative unsupervised feature learning with convolutional
  neural networks.
\newblock In {\em NIPS}, 2014.

\bibitem{fe2003bayesian}
L.~Fei-Fei, R.~Fergus, and P.~Perona.
\newblock A bayesian approach to unsupervised one-shot learning of object
  categories.
\newblock In {\em ICCV}, 2003.

\bibitem{finn2017model}
C.~Finn, P.~Abbeel, and S.~Levine.
\newblock Model-agnostic meta-learning for fast adaptation of deep networks.
\newblock In {\em ICML}, 2017.

\bibitem{graves2013hybrid}
A.~Graves, N.~Jaitly, and A.-r. Mohamed.
\newblock Hybrid speech recognition with deep bidirectional lstm.
\newblock In {\em ASRU}, 2013.

\bibitem{hariharan2016low}
B.~Hariharan and R.~Girshick.
\newblock Low-shot visual object recognition.
\newblock In {\em ICCV}, 2017.

\bibitem{he2016deep}
K.~He, X.~Zhang, S.~Ren, and J.~Sun.
\newblock Deep residual learning for image recognition.
\newblock In {\em CVPR}, 2016.

\bibitem{Hochreiter:NC97}
S.~Hochreiter and J.~Schmidhuber.
\newblock Long short-term memory.
\newblock {\em Neural Computation}, 1997.

\bibitem{kaiser2017learning}
{\L}.~Kaiser, O.~Nachum, A.~Roy, and S.~Bengio.
\newblock Learning to remember rare events.
\newblock {\em ICLR}, 2017.

\bibitem{kingma2014adam}
D.~Kingma and J.~Ba.
\newblock Adam: A method for stochastic optimization.
\newblock In {\em ICLR}, 2015.

\bibitem{koch2015siamese}
G.~Koch, R.~Zemel, and R.~Salakhutdinov.
\newblock Siamese neural networks for one-shot image recognition.
\newblock In {\em ICML Workshop on Deep Learning}, 2015.

\bibitem{lake2015human}
B.~M. Lake, R.~Salakhutdinov, and J.~B. Tenenbaum.
\newblock Human-level concept learning through probabilistic program induction.
\newblock {\em Science}, 2015.

\bibitem{lei2015predicting}
J.~Lei~Ba, K.~Swersky, S.~Fidler, et~al.
\newblock Predicting deep zero-shot convolutional neural networks using textual
  descriptions.
\newblock In {\em ICCV}, 2015.

\bibitem{maaten2008visualizing}
L.~v.~d. Maaten and G.~Hinton.
\newblock Visualizing data using {t-SNE}.
\newblock {\em JMLR}, 2008.

\bibitem{miller2016key}
A.~Miller, A.~Fisch, J.~Dodge, A.-H. Karimi, A.~Bordes, and J.~Weston.
\newblock Key-value memory networks for directly reading documents.
\newblock In {\em EMNLP}, 2016.

\bibitem{munkhdalai2017meta}
T.~Munkhdalai and H.~Yu.
\newblock Meta networks.
\newblock In {\em ICML}, 2017.

\bibitem{noh2016image}
H.~Noh, P.~Hongsuck~Seo, and B.~Han.
\newblock Image question answering using convolutional neural network with
  dynamic parameter prediction.
\newblock In {\em CVPR}, 2016.

\bibitem{pan2016jointly}
Y.~Pan, T.~Mei, T.~Yao, H.~Li, and Y.~Rui.
\newblock Jointly modeling embedding and translation to bridge video and
  language.
\newblock In {\em CVPR}, 2016.

\bibitem{pan2017to}
Y.~Pan, Z.~Qiu, T.~Yao, H.~Li, and T.~Mei.
\newblock To create what you tell: Generating videos from captions.
\newblock In {\em MM}, 2017.

\bibitem{ravi2017optimization}
S.~Ravi and H.~Larochelle.
\newblock Optimization as a model for few-shot learning.
\newblock In {\em ICLR}, 2017.

\bibitem{russakovsky2015imagenet}
O.~Russakovsky, J.~Deng, H.~Su, J.~Krause, S.~Satheesh, S.~Ma, Z.~Huang,
  A.~Karpathy, A.~Khosla, M.~Bernstein, et~al.
\newblock Imagenet large scale visual recognition challenge.
\newblock {\em IJCV}, 2015.

\bibitem{santoro2016meta}
A.~Santoro, S.~Bartunov, M.~Botvinick, D.~Wierstra, and T.~Lillicrap.
\newblock Meta-learning with memory-augmented neural networks.
\newblock In {\em ICML}, 2016.

\bibitem{schmidhuber1992learning}
J.~Schmidhuber.
\newblock Learning to control fast-weight memories: An alternative to dynamic
  recurrent networks.
\newblock {\em Neural Computation}, 1992.

\bibitem{schuster1997bidirectional}
M.~Schuster and K.~K. Paliwal.
\newblock Bidirectional recurrent neural networks.
\newblock {\em IEEE Transactions on Signal Processing}, 1997.

\bibitem{sukhbaatar2015end}
S.~Sukhbaatar, J.~Weston, R.~Fergus, et~al.
\newblock End-to-end memory networks.
\newblock In {\em NIPS}, 2015.

\bibitem{vinyals2016matching}
O.~Vinyals, C.~Blundell, T.~Lillicrap, D.~Wierstra, et~al.
\newblock Matching networks for one shot learning.
\newblock In {\em NIPS}, 2016.

\bibitem{wang2016learning}
Y.-X. Wang and M.~Hebert.
\newblock Learning to learn: Model regression networks for easy small sample
  learning.
\newblock In {\em ECCV}, 2016.

\bibitem{weston2014memory}
J.~Weston, S.~Chopra, and A.~Bordes.
\newblock Memory networks.
\newblock In {\em ICLR}, 2014.

\bibitem{yao2017boosting}
T.~Yao, Y.~Pan, Y.~Li, Z.~Qiu, and T.~Mei.
\newblock Boosting image captioning with attributes.
\newblock In {\em ICCV}, 2017.

\bibitem{Yosinski:NIPS14}
J.~Yosinski, J.~Clune, Y.~Bengio, and H.~Lipson.
\newblock How transferable are features in deep neural networks?
\newblock In {\em NIPS}, 2014.

\end{thebibliography}
}

\end{document}